\documentclass[10pt,twocolumn,letterpaper]{article}

\usepackage{times}
\usepackage{epsfig}
\usepackage{graphicx}
\usepackage{amsmath}
\usepackage{amssymb}

\usepackage{epstopdf}
\DeclareMathOperator{\Tr}{tr} 
\DeclareMathOperator{\logdet}{logdet} 
\usepackage{enumitem}
\usepackage{amsthm}

\usepackage[linesnumbered,ruled,vlined]{algorithm2e}
\usepackage[stable]{footmisc}
\usepackage{floatrow}

\begin{document}

%%%%%%%%% TITLE
%\title{Euclidean Space Region Covariance Representation Sparse Coding for Image Classification}
\title{Parameterizing Region Covariance: An Efficient Way To Apply Sparse Codes On Second Order Statistics}

% Authors at the same institution
%\author{First Author \hspace{2cm} Second Author \\
% Authors at different institutions
\author{Xiyang Dai \\
University of Maryland\\
{\tt\small xdai@cs.umd.edu}
\and
Sameh Khamis \\
	University of Maryland\\
	{\tt\small sameh@umiacs.umd.edu}
\and
Yangmuzi Zhang \\
	University of Maryland\\
	{\tt\small ymzhang@umiacs.umd.edu}
\and
Larry S. Davis \\
	University of Maryland\\
	{\tt\small lsd@umiacs.umd.edu}	
}

\maketitle

%%%%%%%%% ABSTRACT
\begin{abstract}
Sparse representations have been successfully applied to signal processing, computer vision and machine learning. Currently there is a trend to learn sparse models directly on structure data, such as region covariance. However, such methods when combined with region covariance often require complex computation. We present an approach to transform a structured sparse model learning problem to a traditional vectorized sparse modeling problem by constructing a Euclidean space representation for region covariance matrices. Our new representation has multiple advantages. Experiments on several vision tasks demonstrate competitive performance with the state-of-the-art methods. 
\end{abstract}

%%%%%%%%% BODY TEXT
\section{Introduction}
Sparse representations have been successfully applied to many tasks in signal processing, computer vision and machine learning. Many algorithms\cite{KSVD, LCKSVD} have been proposed to learn an over-complete and reconstructive dictionary based on such representations. These algorithms involve vectorizing the input data which can destroy inherent ordering information in the data\cite{VecNotWork1, TSC}. Instead sparse codes can be constructed directly based on the original structure of the input data. Such structures include diffusion tensors, region covariance, etc. The region covariance structure, introduced by Tuzel \textit{et al.}\cite{RegCov} provides a natural way to fuse different features for a given region. Additionally, the averaging filter in covariance computation reduces noise that corrupts individual samples. Furthermore, Porikli \textit{et al.}\cite{FastRegCov} showed that it can be constructed for arbitrary-sized windows in constant time using integral images. Hence, it has become a popular descriptor for  face recognition\cite{GaborFace, FaceAAM, SDL}, human detection\cite{HumanD}, tracking\cite{HumanD}, object detection \cite{RSR, TSCwD}, action recognition \cite{logE-SR, ActionR} and pedestrian detection \cite{PedD}. 

However, region covariance matrices are positive definite matrices, forming a connected Riemannian manifold. Current manifold-based methods for region covariance often require complex computation. Many applications remain restricted to k-nearest-neighbors or kernel SVMs, using geodesic distance measurement\cite{RFramework, HumanD, PedD}. Pennec \textit{et al.} \cite{RFramework} first introduced the general framework to calculate the statistics based on an affine-invariant metric. Recently, there have been several attempts to develop sparse coding for region covariance matrices\cite{ActionR, RSR,TSC,LogDet,TSCwD,SDL}. However, such approaches all involve complex computations, including calculating eigenvalues, matrix logarithms and matrix determinants.

\begin{figure*}
	\begin{center}
		\includegraphics[scale = 0.3]{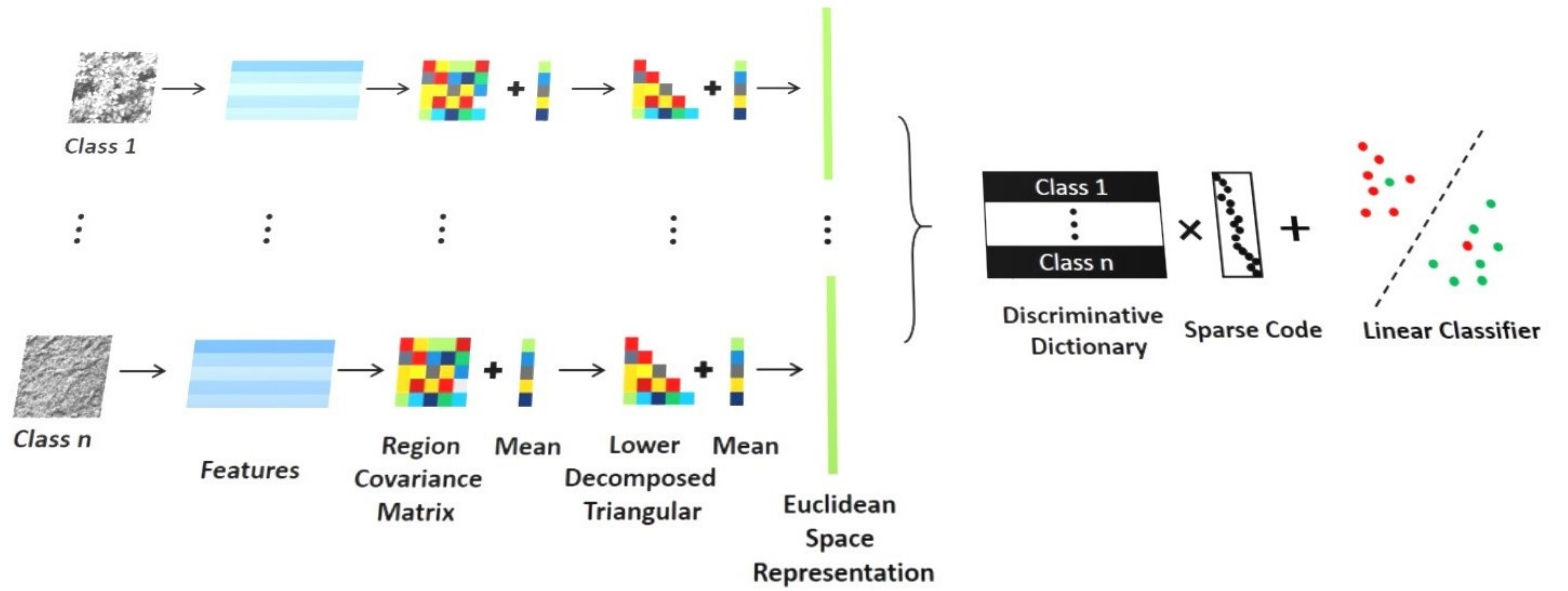}
	\end{center}
	\caption{The framework of our approach when applied to texture classification problem.}
	\label{framework_fig}
\end{figure*}

We present an approach for sparse coding parameterized representations of region covariance matrices inspired by finance applications. This representation preserves the same second order statistics as region covariance matrices. More importantly, the representation is Euclidean and hence can be vectorized and computed effectively in the traditional sparse coding framework. We further learn discriminative dictionaries over this representation by integrating label consistency regularization and class information into the objective function. The framework of our approach is shown in Figure \ref{framework_fig}. The main contributions of this paper are:
\begin{itemize}[noitemsep]
	\item Introduction of covariance parameterization used in finance to the computer vision community.
	\item Design of a new Euclidean representation for region covariance that has multiple advantages, including lower time complexity for measuring similarity and preserving both first order and second order statistics of a given region.
	\item Performing discriminative dictionary learning on our new representation of region covariance to show its effectiveness.
	\item Experiments show state-of-the-art performance on multiple tasks.
\end{itemize}
\section{Background}
We provide a brief review of the region covariance descriptor and its corresponding similarity measurement methods. 
\subsection{Region Covariance Descriptors}
Given an image $I$, let $\varPhi$ be a function that extracts a $d$-dimension feature vector $z_i$ at each pixel $i \in I$, i.e. $\varPhi(I,x_i,y_i) = z_i$, where $z_i \in R^d$ and $(x_i, y_i)$ is the location of pixel $i$. $\varPhi$ can be any feature mapping function such as intensity, gradient, different color channels, filter responses, etc. $F$ is a $W \times H \times d$ dimensional feature matrix extracted from $I$. A given image region $R$ is represented by the $d \times d$ covariance matrix $C_R$ of the set of feature vectors $\{z_i\}_{i=1}^N$ of all $N$ points inside the region $R$. The region covariance descriptor $C_R \in R^{d \times d}$ is defined as:

\begin{equation}
	C_R = \frac{1}{N-1}\sum_{i=1}^{N} (z_i - \mu_R)(z_i - \mu_R)^T
\end{equation}
where, $\mu_R$ is the mean vector,
\begin{equation}
	\mu_R = \frac{1}{|N|}\sum_{i=1}^{|N|} z_i
\end{equation}

\subsection{Positive Definite Similarity Computation}
In general, covariance matrices are positive definite, except for some special cases. They are usually regularized to make them strictly positive definite. Hence, the region covariance descriptors belong to the $d \times d$ positive definite space $S_{++}^d$, which lies on a Riemannian manifold, not in Euclidean space. This fact makes the similarity measurement between two covariance matrices non-trivial. One well-known method for computing similarity is the Affine Invariant Riemannian Metric (AIRM)\cite{RFramework} which uses the corresponding geodesic distance on the manifold as a similarity measurement:
\begin{equation}
	D_R(X,Y) = ||\log(X^{-1/2}YX^{-1/2})||_F
\end{equation}
where $\log(\cdot)$ is the matrix logarithm and $||\cdot||_F$ is the Frobenius norm. This method is widely used in classification tasks that involve region covariance. However, the requirement of eigenvalue computation makes it very expensive when used in iterative optimization frameworks. 

Many methods have been proposed to improve AIRM. One is the Log-Euclidean Riemannian Metric (LERM)\cite{LE}:
\begin{equation}
	D_{le}(X,Y) = ||\log(X)-\log(Y)||_F
\end{equation}
This method maps the positive definite matrices into a flat Rieminnian space by taking the logarithm of the matrices so that the Euclidean distance measurement can be used. While the logarithm for each of these matrices can be evaluated offline, computing the matrix logarithm is still expensive.     

More recently, LogDet divergence\cite{LogDet} has been investigated:
\begin{equation}
	D_{ld}(X,Y) = \Tr(XY^{-1}) - \logdet(XY^{-1}) - n
\end{equation}
where $\logdet(\cdot)$ is the logarithm of a matrix determinant and $\Tr(\cdot)$ is the  matrix trace. This method was used in several tensor based sparse coding methods\cite{TSC,TSCwD,SDL}. The LogDet divergence reduces computational complexity by replacing the calculation of eigenvalues with determinants. Also, it avoids the explicit manifold embedding and results in a convex MAXDET problem. However, since the computation of matrix determinants each iteration is still roughly $O(d^3)$, where $d$ is the column size of the region covariance matrix, the whole optimization process is still costly.

\section{A Euclidean Space Representation for Region Covariance}
In this section, we introduce our methods to construct a small set of points that lie in Eucliddean space and preserve the second order statistics.
\subsection{Understanding the Region Covariance}
Covariance matrices used in finance usually represent the variance of stock price and the correlations between different stocks. Region covariance in computer vision applications shares similar concepts. Given a set of features $Z = [f^1, f^2, \cdots f^n]$, for all $N$ points in a region, the region covariance can be written as:
%\begin{equation}
%	\left\{
%	\begin{array}{c c c c}
%		\sum^{N}_{i=1} (f^1_i - \mu_{f^1})^2 & <f^1 - \mu_{f^1}, f^2 - \mu_{f^2} > & \cdots\ & <f^1 - \mu_{f^1}, f^n - \mu_{f^n} > \\
%		<f^2 - \mu_{f^2}, f^1 - \mu_{f^1} >& \sum^{N}_{i=1} (f^2_i - \mu_{f^2})^2 & \cdots & <f^2 - \mu_{f^2}, f^n - \mu_{f^n} > \\
%		\vdots & \vdots & \ddots & \vdots\\
%		<f^n - \mu_{f^n}, f^1 - \mu_{f^1} > & <f^n - \mu_{f^n}, f^2 - \mu_{f^2} > & \cdots & \sum^{N}_{i=1} (f^n_i - \mu_{f^n})^2\\
%	\end{array}
%	\right\}
%\end{equation}

\begin{equation}
	\left\{
	\scalebox{0.8}{	
	\begin{tabular}{c c c}
		$\sum^{N}_{i=1} (f^1_i - \mu_{f^1})^2$  & $\cdots$ & $<f^1 - \mu_{f^1}, f^n - \mu_{f^n} >$ \\
		$\vdots$ & $\ddots$ & $\vdots$\\
		$<f^n - \mu_{f^n}, f^1 - \mu_{f^1} >$  & $\cdots$ & $\sum^{N}_{i=1} (f^n_i - \mu_{f^n})^2$\\
	\end{tabular}}
	\right\}	
\end{equation}
where $f^n_i$ is the $n$th feature value for point $i$ and $\mu_{f^n}$ is the mean of the $n$th feature vector. The diagonal entries of the covariance matrix represent the variances of each feature, while the entries outside the diagonal represent the correlations of different features. To design a covariance representation, we want to include both of these terms.
\subsection{Cholesky Decomposition}
A meaningful region covariance matrix $C$ should be symmetric and positive semidefinite, and hence can be decomposed as the product:
\begin{equation}
C = LL^T
\end{equation}
A obvious way to calculate $L$ is using Cholesky decomposition, which enjoys low computation cost and preserves some properties of the covariance matrix\cite{sigma_set}. 
%Such a method has been explored by Hong \textit{et al.} \cite{sigma_set} and applied to vision tasks. %
Let $L_x, L_y$ be the lower triangular matrices calculated from $C_x, C_y$ using Cholesky decomposition, the distance between $C_x, C_y$ can be approximated by
%Let $C_x$ and $C_y$ be two covariance matrices, we first use Cholesky decomposition to generate the lower triangular matrices $L_x, L_y$ from $C_x, C_y$:
%\begin{equation}
%	C_X = L_XL_X^T,\quad	C_Y = L_YL_Y^T 
%\end{equation}
%Then, the distance between $C_X, C_Y$ can be calculated in Euclidean space by: 
\begin{equation}
	D_{chol}(C_X,C_Y) = ||L_Xe - L_Ye||_F
\end{equation}
where $e$ is a standard Euclidean basis. The Cholesky decomposition guarantees that the new representation $s = Le$ is unique for any covariance matrix $C$. 

Although the representation based on Cholesky decomposition works in practice, it is difficult to interpret the entries in the lower triangular matrix. In particular, it is difficult to obtain the correlation coefficients which are available in the original covariance matrix.
\subsection{Spherical Decomposition}
Alternatively, we seek a lower triangular representation that not only obeys the decomposition rule, but also possesses better statistical interpretations. Inspired by the spherical parametrization method used in finance application\cite{sphere_parameterization} for covariance estimation, a new representation can be constructed using spherical coordinates, which involves a series of rotational mappings from the standard basis to the lower triangular matrix\cite{parameterization}. We start with the lower triangular matrix $L$ generated from Cholesky decomposition, and then represent it as: %Instead, an alternative approach is to directly model the manifold structure. If we approximate the manifold using a sphere surface, we can represent the covariance matrix in spherical coordinates: % 

\begin{equation}
	L_{i,j} = \left\{ 
	\begin{array}{l l}
		\tilde{s}_{i,1} \cos(\tilde{s}_{i,2}) & \quad \text{$j$=1} \\
		\tilde{s}_{i,1} \cos(\tilde{s}_{i,j+1}) \Pi_{k=2}^{j} \sin(\tilde{s}_{i,k}) & \quad \text{$2\le j \le i-1$} \\
		\tilde{s}_{i,1} \Pi_{k=2}^{i} \sin(\tilde{s}_{i,k}) & \quad \text{$j=i$} \\
		0 & \quad \text{$i+1\le j\le n$}  	
	\end{array}\right.
	\label{eq:sphere}
\end{equation}
where $\tilde{s}_{i,j}$ denotes the new representation, $L_{i,j}$ is an element of $L$. A special case of \ref{eq:sphere} is $L_{1,1} = \tilde{s}_{1,1}$. To ensure the uniqueness of converting from a covariance matrix to spherical coordinates, we must have:
\begin{equation}
	\left\{ 
	\begin{array}{l l}
		\tilde{s}_{i,1} > 0, &\quad i = 1,\cdots,n \\
		\tilde{s}_{i,j} \in (0,\pi) &\quad i = 2,\cdots,n \quad j = 2,\cdots,i
	\end{array}\right.
\end{equation} 
This new representation has the following statistical advantages:
\begin{itemize}[noitemsep]
	\item The diagonal entries of the covariance matrix are captured directly by the entries of this new representation: $C_{i,i} = \tilde{s}_{i,1}^2$.
	\item Some of the correlation coefficients $\rho$ for the covariance matrix can be uniquely mapped to the new representation: $\rho_{1,i} = \cos(\tilde{s}_{i,2}), \quad i = 2,\cdots, n$.
	\item Elements of the new representation are independent of each other.
\end{itemize}
The new representation lies in Cartesian space \cite{parameterization}, hence the distance can be measured using the Frobenius norm:
\begin{equation}
	D_{sphere}(C_X,C_Y) = ||\tilde{s}_X - \tilde{s}_Y||_F
\end{equation}
where $\tilde{s}_X$ and $\tilde{s}_Y$ are the new representations calculated by \ref{eq:sphere}.
\subsection{Combine with the Mean}
The mean of the original features can be concatenated to $s$ to make it more informative and robust:
\begin{equation}
	s = \lambda\mu_R \bigcup s
\end{equation}
$\lambda$ is a parameter that balances the scale difference between the mean and our representation. %For Cholesky decomposition the proper scale is $\lambda = \frac{1}{\sqrt{d}}$, where $d$ is the dimension of the covariance matrix. For spherical decomposition the proper scale is $\lambda = \frac{\pi}{\sqrt{d}}$. %

Our representation $s$ lies in Euclidean space and the similarity between representations can be simply measured by the Frobenius norm. Compared to the traditional covariance matrix, our new representation enjoys several advantages:
\begin{itemize}[noitemsep]
	%\item Lower space complexity. The space complexity of our representation is $\frac{d^2}{2}+\frac{3d}{2}$ compared to the original $d^2$ covariance matrix.
	\item Lower time complexity for measuring similarity. The time complexity of using the Frobenius norm to measure the similarity is $O(d^2)$ compared to $O(d^3)$ of AIRM\cite{RFramework} and LogDet\cite{LogDet}. 
	\item Informative and robust. Our new representation preserves both the first and the second order statistics. Since the region covariance only captures the differences between features, it may lose some useful statistics within separate feature channels. Hence, fusing feature means into our representation enhances robustness.
	\item Flexibility. The similarity measurement of our new representation can be calculated in Euclidean space, which enables applying many traditional learning methods to second order statistic.
\end{itemize}
\section{Discriminative Sparse Coding and Dictionary Learning}
We next describe a method to learn a reconstructive and discriminative dictionary from multi-class data. We construct a sub-dictionary for each class. We explicitly encourage independence between dictionary atoms from different sub-dictionaries and leverage class information in the optimization problem. We adopt the LC-KSVD\cite{LCKSVD} method to learn the discriminative dictionary.
\subsection{Dictionary Learning via Label Consistent Regularization}
Let S be a set of N d-dimensional Euclidean space region covariance representations for training dictionary, i.e. $S = [s_1, s_2, \dots s_N]\in R^{d\times N}$. Learning a reconstructive dictionary with K atoms for sparse representation of $S$ can be formulated as:
\begin{align}
	\label{final_eq}
	\arg\min_{D,X,A,W}&||S-DX||_F^2 + \alpha||Q-AX||_F^2 + \beta||H-WX||_F^2 \nonumber\\
	s.t.& \quad \forall i, ||x_i||_0\le T 
\end{align}
where $D \in R^{d \times K}, K\ge d$ is the learned over-complete dictionary. $X \in R^{K\times N}$ is the sparse codes for given inputs, $Q\in R^{K \times N}$ are the "discriminative" sparse codes, $A$ is a linear transformation matrix defined to transform the original sparse codes $X$ to be most discriminative in sparse feature space, $W$ denotes the parameters of a linear classifier $f(x;W) = Wx$, $H\in R^{m\times N}$ are the class labels and $T$ is a sparsity constraint factor. 

Minimizing the objective function not only encourages independence between dictionary atoms from different sub-dictionaries, but also trains a linear classifier simultaneously. We use the efficient K-SVD algorithm to find the optimal solution for all parameters simultaneously. 
\subsection{Classification}
After obtaining the dictionary $D$ and the linear classifier parameter $W$, the sparse representation $X_{test}$ for the test inputs $S_{test}$ can be calculated as:
\begin{align}
	\label{sparse_code}
	X_{test} = \arg\min_{X}& \quad ||S_{test}-DX||_F^2 \nonumber\\
	s.t.& \quad \forall i, ||x_i||_0\le T 
\end{align}
We simply use the linear classifier $f(x;W) = Wx$ to estimate the label of a test sample $x_i \in X_{test}$:
\begin{align}
	l = \arg\max_{l} (Wx_i)
\end{align}
\subsection{Sparse Codes as Features}
We can also fuse the generated sparse codes with other features. One drawback of \ref{sparse_code} is that the sparsity constraint factor $T$ is a hard threshold that forces the sparse codes to have fewer than T non-zero items. This is good for classification task, but when using sparse codes as features we are not concerned with the number of non-zero items. Instead we want to make the sparse codes as informative as possible. Hence, we consider a "soft" version of equation \ref{sparse_code}:
\begin{align}
	X_{test} = \arg\min_{X}& \quad ||S_{test}-DX||_F^2 + t_1||X||_1 + \frac{t_2}{2}||X||^2_2
\end{align}
where $t_1$ and $t_2$ are the new sparsity constraint factors. These two parameter control the generation of more continuous sparse codes. 

\section{Experiments}
\begin{figure}[h]
	\includegraphics[scale=0.15]{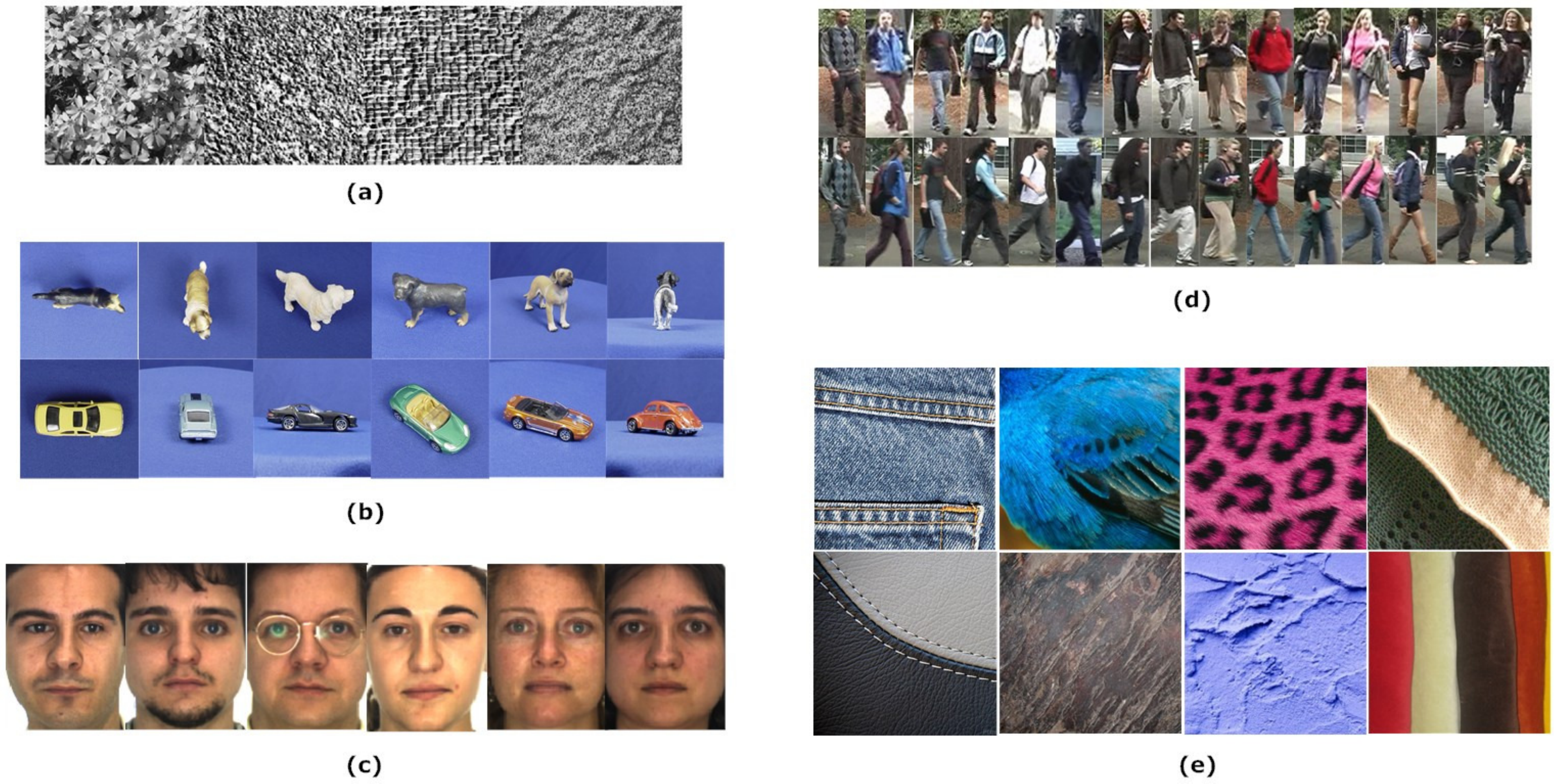}
	\caption{Sample images from experiment datasets for different tasks: (a) Texture classification, (b) Object classification, (c) Face recognition, (d) Person re-identification and (e) Material classification}
	\label{sample}	
\end{figure}

We evaluate our approach on several different tasks: texture classification, object classification, face recognition, material classification and person re-identification. Sample images for different tasks are shown in Figure \ref{sample}.
For fair comparison, we experiment on the same features as reported by other methods.  

\subsection{Texture Classification}

\begin{table*}
	\begin{center}
		\begin{tabular}{|l|c|c|c|c|c|c|}
			\hline
			Scenario & logE-SR\cite{logE-SR} & TSC\cite{TSCwD} & RSR\cite{RSR} &  SDL\cite{SDL} & \textbf{Ours-chol} & \textbf{Ours-sphere} \\
			\hline\hline
			5c & 0.88&	\textbf{1.00}&	0.99&	0.99 & 	0.99 	& 0.98\\
			5m & 0.54&	0.73&	0.85&	0.95 & 	\textbf{0.96} 	& \textbf{0.97}\\
			5v & 0.73&	0.86&	0.89&	0.90 &	\textbf{0.92}	& \textbf{0.91}\\
			5v2 & 0.70&	0.85&	0.89&	0.93 & 	\textbf{0.94}	& \textbf{0.95}\\
			5v3 & 0.65&	0.83&	0.87&	0.84 & 	\textbf{0.97}	& \textbf{0.92}\\
			10 & 0.60&	0.81&	0.85&	\textbf{0.89} & 	0.82	& 0.81\\
			10v & 0.64&	0.68&	0.86&	\textbf{0.91} &	0.84	& 0.79\\
			16c &  0.68&	0.75&	0.83&	0.86 &	\textbf{0.89}	& \textbf{0.89}\\
			16v & 0.56&	0.66&	0.77&	\textbf{0.89} &	0.79	& 0.79\\
			\hline 
		\end{tabular}
	\end{center}
	\caption{Texture classification results on the Brodatz dataset.}
	\label{texture_table}
\end{table*}

\textbf{Evaluation Protocol.} We follow the protocol in \cite{BrodatzDataset} to create mosaics under nine test scenarios. Each scenario has various numbers of classes, including 5-textures, 10-textures and 16-textures. Each image in the dataset is resized to $256 \times 256$ and cut into $32 \times 32$ non-overlapped blocks, yielding 64 data samples per image. For each scenario, we randomly select 5 data samples as training and use the rest for testing. The evaluation is repeated 10 times.

\textbf{Implementation Details.}  We extract features $F(x) = \big\{I, |I_x|, |I_y|, |I_{xx}|, |I_{yy}|\big\}$ based on intensity and gradient from each sample. They form a $5 \times 5$ region covariance matrix and result in a 20-dimension vector in our representation. We use the same parameter configuration ($\sqrt{\alpha} = 5, \sqrt{\beta} = 5, T = 1$) in all test scenarios. 

\textbf{Results.} Table \ref{texture_table} shows the classification results under the nine scenarios. We compare our method with logE-SR\cite{logE-SR}, TSC\cite{TSCwD}, RSR\cite{RSR}, SDL\cite{SDL}. The mean accuracy of our method achieves the best result in over half of the scenarios (5m, 5v, 5v2, 5v3, 16c). Overall, our maximum classification results over 10 runs are comparable to the best scores.

\subsection{Object Classification}
\textbf{Evaluation Protocol.} The ETH80 dataset\cite{eth80} contains eight objects with ten instances each collected form 41 different views. There are 3280 images total. Images for each object have large view point changes which make this dataset very challenging for object recognition task. 

\textbf{Implementation Details.}
For each image, we generate a $19\times19$ covariance matrix with feature $F = \{x, y, R, G, B, |I_x|, |I_x|, |I_{LoG}|, \sqrt{I_x^2 + I_y^2}, F_{Laws}\}$, where $I_{Log}$ is the responses from Laplacian of Gaussian filter, $ F_{Laws}$ is the responses form the bank of Laws texture filters \cite{laws};

\textbf{Results.}
We randomly split 80\% of dataset for training and use the rest for testing. We repeat the procedure 10 times and report the average accuracy. We compare our results with several state-of-the-art methods \cite{TSCwD, kle, RSR, rsc}, shown in Table \ref{object_table}. Our results are comparable to other methods.

\begin{table}
		\begin{tabular}{|c|c|}
			\hline
			Method & Accuracy \\		
			\hline\hline
			TSC\cite{TSCwD} & 37.1\\
			K-LE-SC\cite{kle} & 76.6\\
			RSR\cite{RSR} & 81.6\\
			Riem-SC\cite{rsc} & 77.9\\
			\textbf{Ours-chol} & 79.8\\
			\textbf{Ours-sphere} &\textbf{84.0}\\
			\hline
		\end{tabular}
		\caption{Object classification results on the ETH80 dataset.}
		\label{object_table}
\end{table}

\subsection{Face Recognition}
\textbf{Evaluation Protocol.} The AR face dataset\cite{ARDataset} contains over 4000 face images captured from 126 individuals. For each individual, there are 26 images separated in two sessions.  We follow the protocol used in \cite{SDL}, randomly select 10 subjects to evaluate in our experiment. We repeat the evaluation 20 times.

\textbf{Implementation Details.} Each image is cropped to $27 \times 20$ and converted into gray scale. We extract the intensity and the spatial information along with the responses of Gabor filters with 8 orientations $\theta_u = \frac{\pi u}{8}, u\in\{0,1,2,\dots,7\}$ and 5 scales $v\in\{0,1,\dots,4\}$: $F(x,y)=\{I, x, y, |G_{0,0}(x,y)|, \dots, |G_{0,4}(x,y)|, \dots, |G_{7,4}(x,y)|\}$ where $G_{u,v}$ is the response of a 2D Gabor wavelet\cite{Gabor} defined by:
\begin{align}
	G_{u,v} = & \frac{k_v^2}{4\pi^2}\sum_{t,s}\mathrm{e}^{-\frac{k_v^2}{8\pi^2}((x-s)^2+(y-t)^2)}\nonumber\\
	& (\mathrm{e}^{ik_v(x-t)\cos(\theta_u)+(y-s)\sin(\theta_u)}-\mathrm{e}^{-2\pi^2})
\end{align}
where $k_v = \frac{1}{\sqrt{2^{v-1}}}$.

\textbf{Results.}
We randomly select 15, 18, 21 images per person to train and use the rest for testing. Our results are compared with \cite{TSCwD,RSR,SDL}. Table \ref{face_table} contains our results. We achieve significant performance improvements on all three test configurations.
\begin{table*}
	\begin{center}
		\begin{tabular}{|l|c|c|c|c|c|c|c|}
			\hline
			$\#$train sample &  TSC\cite{TSCwD} & RSR\cite{RSR} & SDL\cite{SDL} & \textbf{Ours-chol} & \textbf{Ours-sphere}\\
			\hline\hline
			15 per person& 	78.6&	81.4& 82.3&  \textbf{86.7}&  \textbf{89.1}\\
			18 per person& 	79.9&	84.1& 85.2&  \textbf{89.0}&  \textbf{90.2}\\
			21 per person& 	80.8&	85.7& 86.1&  \textbf{90.6}&  \textbf{91.8}\\
			\hline
		\end{tabular}
	\end{center}
	\caption{Face recognition results on the AR face dataset.}
	\label{face_table}
\end{table*}

\subsection{Material Classification}
\textbf{Evaluation Protocol.} The UIUC material dataset\cite{SD} contains eighteen categories with twelve images each (mainly belong to bark, fabric, construction materials, outer coat of animals and so on). Images for each category have different scales and are collected in the wild, which make this dataset very difficult. This dataset is considered as one of the-state-of-art benchmarks for material classification task. The standard evaluation protocol is to randomly split half of dataset for training and use the rest for testing. We report the average accuracy over 10 repeats. 

\textbf{Implementation Details.}
For each image, we generate a $155\times155$ covariance matrix usng 128 dimensional SIFT feature and 27 color feature ($3\times3$ raw RGB pixels around the center of SIFT descriptor). We calculate above region covariance matrices over a $12\times12$ window size with a 4 step size.
\begin{table}
	\begin{tabular}{|l|c|}
		\hline
		Method & Accuracy \\
		\hline\hline
		SD\cite{SD} & 43.5\%\\
		CDL\cite{CDL} & 52.3\%\\		
		RSR\cite{RSR} & 52.8\%\\
		\textbf{Ours-chol} & \textbf{57.2}\%\\
		\textbf{Ours-sphere} & \textbf{56.8}\%\\	
		\hline
	\end{tabular}
	\caption{Material classification results on the UIUC material dataset.}	
	\label{material_table}
\end{table}

\textbf{Results.}
We compare our results with several state-of-the-art methods \cite{SD, CDL, RSR}, shown in Table \ref{material_table}. Our results are comparable to other methods.   

\subsection{Person Re-identification}
\textbf{Evaluation Protocol.} The VIPeR dataset\cite{VIPeRDataset} contains 632 pedestrian pairs captured from different camera views. Each image in the pair is resized to $128 \times 48$. They exhibit large viewpoint variations among pedestrian pairs, which makes it one of the most challenging datasets in person re-identification. We follow the protocol widely used in \cite{SDALF}, splitting the 632 pedestrian pairs into half for training and half for testing. Two-fold validation is applied during evaluation. We repeat the evaluation 10 times and report the average result.

\textbf{Implementation Details.} We extract $9 \times 9$ blocks with a stride of 4 from each image. For each block, we extract gradient and color features in different channels (including RGB, HSV and color name\cite{ColorName}) $F(x) = \big\{I, |I_x|, |I_y|, R, G, B, H, S, V, cln\big\}$ to form region covariance matrices. This generates a $10 \times 10$ region covariance matrix and result in a 65-dimensional vector in our representation. We then learn sparse codes and use them as features. Additionally, we also extract color histograms from different channels (Lab, HSV and color name\cite{ColorName}) using $7 \times 48$ stripes with a stripe of 3 for consistency with our region covariance sparse code feature. The color histograms are further reduced to 300 dimensions by PCA. We concatenate these two features together with normalizing the maximum value to 1 for each sample and use information theoretic metric learning method\cite{itml} to learn the final ranks.

\begin{figure}

			\includegraphics[scale =0.29]{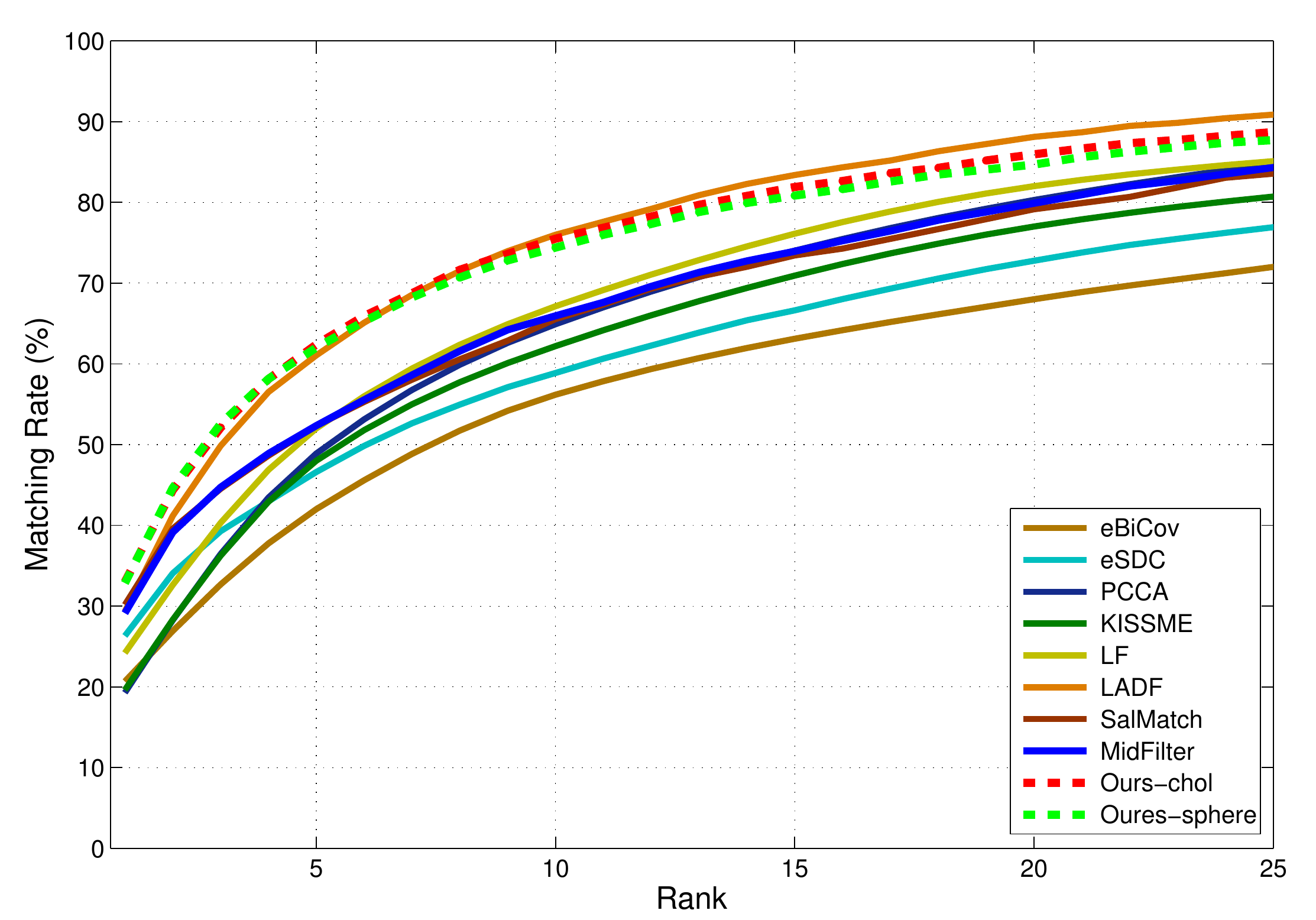}

		\caption{CMC curves on the VIPeR dataset.}
		\label{reid_fig}
\end{figure}

\begin{figure}
	\begin{center}
		\includegraphics[scale = 0.34]{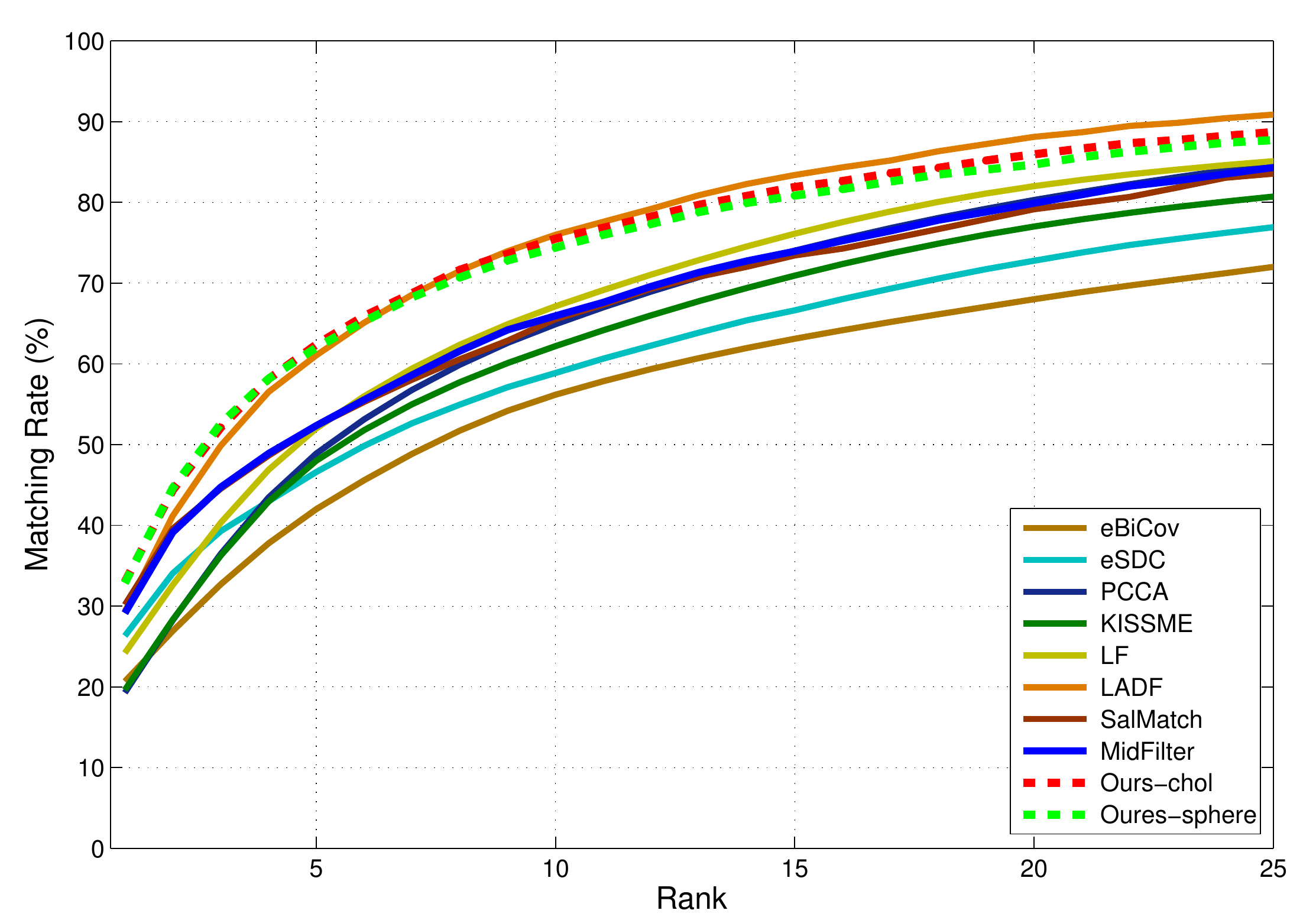}
	\end{center}
	\caption{A Example of the matching pairs at each rank on VIPeR dataset. Green box shows the probe images, red box shows the ground truth matches. Each row shows the retrieved results from top matches.}
	\label{reid_vslz_fig}
\end{figure}

\begin{table}
		\begin{tabular}{|l|c|}
			\hline
			Method & Rank 1 Accuracy \\
			\hline\hline
			eBiCov\cite{eBiCov} & 20.66\\
			eSDC\cite{eSDC} & 26.31\\		
			PCCA\cite{PCCA} & 19.27\\
			KISSME\cite{KISSME} & 19.60\\
			LF\cite{LF} & 24.18\\
			LADF\cite{LADF} & 29.34\\
			SalMatch\cite{SalMatch} & 30.16\\
			MidFilter\cite{MidFilter} & 29.11\\
			\textbf{Ours-chol} &\textbf{32.99}\\
			\textbf{Ours-sphere} &\textbf{32.84}\\	
			\hline
		\end{tabular}
		\caption{Rank 1 matching accuracy on the VIPeR dataset.}%
		\label{reid_table}
\end{table}

\textbf{Results.} We compare our method with state-of-the-art methods that don't require foreground priors such as  PCCA\cite{PCCA}, KISSME\cite{KISSME}, eBiCov\cite{eBiCov}, eSDC\cite{eSDC}, LF\cite{LF}, SalMatch\cite{SalMatch}, LADF\cite{LADF} and MidFilter\cite{MidFilter}. Table \ref{reid_table} shows the rank 1 accuracy on the VIPeR dataset. The rank 1 results of our method outperform all the competing methods. Figure \ref{reid_fig} contains the cmc ranking curve from rank 1 to rank 25. Our curve is competitive to most of the state-of-the-arts methods. By visualizing the matching pairs (shown in Figure \ref{reid_vslz_fig}), we find our approach is good at finding discriminative textures thanks to our region covariance representation.

\section{Conclusion}
We introduced a new representation for region covariance which lies in Euclidean space. This new representation not only shares the same second order statistics with covariance matrices, but also includes the first order statistics. Analysis shows its space and computation advantages over region covariance matrices. Additionally, the discriminative dictionary learning problem on this representation can be solved efficiently in the traditional K-SVD framework. Experiments on different tasks demonstrate the proposed approach is effective and robust.

{\small
\bibliographystyle{ieee}
\bibliography{egbib}

\begin{thebibliography}{10}\itemsep=-1pt

\bibitem{KSVD}
M.~Aharon, M.~Elad, and A.~Bruckstein.
\newblock K-svd: An algorithm for designing overcomplete dictionaries for
  sparse representation.
\newblock {\em TSP}, 54(11):4311--4322, 2006.

\bibitem{LE}
V.~Arsigny, P.~Fillard, X.~Pennec, and N.~Ayache.
\newblock Log-{Euclidean} metrics for fast and simple calculus on diffusion
  tensors.
\newblock {\em Magnetic Resonance in Medicine}, 56(2):411--421, 2006.

\bibitem{rsc}
A.~Cherian and S.~Sra.
\newblock Riemannian sparse coding for positive definite matrices.
\newblock In {\em ECCV}, 2014.

\bibitem{itml}
J.~Davis, B.~Kulis, P.~Jain, S.~Sra, and I.~Dhillon.
\newblock Information-theoretic metric learning.
\newblock In {\em ICML}, 2007.

\bibitem{VIPeRDataset}
G.~Doug, B.~Shane, and T.~Hai.
\newblock Evaluating appearance models for recognition, reacquisition, and
  tracking.
\newblock In {\em PETS}, 2007.

\bibitem{SDALF}
M.~Farenzena, L.~Bazzani, A.~Perina, V.~Murino, and M.~Cristani.
\newblock Person re-identification by symmetry-driven accumulation of local
  features.
\newblock In {\em CVPR}, 2010.

\bibitem{ActionR}
K.~Guo, P.~Ishwar, and J.~Konrad.
\newblock Action recognition using sparse representation on covariance
  manifolds of optical flow.
\newblock In {\em AVSS}, 2010.

\bibitem{RSR}
M.~Harandi, C.~Sanderson, R.~Hartley, and B.~Lovell.
\newblock Sparse coding and dictionary learning for symmetric positive definite
  matrices: A kernel approach.
\newblock In {\em ECCV}, 2012.

\bibitem{VecNotWork1}
T.~Hazan, S.~Polak, and A.~Shashua.
\newblock Sparse image coding using a 3d non-negative tensor factorization.
\newblock In {\em ICCV}, 2005.

\bibitem{sigma_set}
X.~Hong, H.~Chang, S.~Shan, X.~Chen, and W.~Gao.
\newblock Sigma set: A small second order statistical region descriptor.
\newblock In {\em CVPR}, pages 1802--1809, June 2009.

\bibitem{FaceAAM}
H.~Huo and J.~Feng.
\newblock Face recognition via aam and multi-features fusion on riemannian
  manifolds.
\newblock In {\em ACCV}, 2009.

\bibitem{LCKSVD}
Z.~Jiang, Z.~Lin, and L.~Davis.
\newblock Learning a discriminative dictionary for sparse coding via label
  consistent k-svd.
\newblock In {\em CVPR}, 2011.

\bibitem{KISSME}
M.~Kostinger, M.~Hirzer, P.~Wohlhart, P.~Roth, and H.~Bischof.
\newblock Large scale metric learning from equivalence constraints.
\newblock In {\em CVPR}, 2012.

\bibitem{LogDet}
B.~Kulis, M.~Sustik, and I.~Dhillon.
\newblock Learning low-rank kernel matrices.
\newblock In {\em ICML}, 2006.

\bibitem{laws}
K.~Laws.
\newblock Rapid texture identification.
\newblock {\em Proc. SPIE}, 0238:376--381, 1980.

\bibitem{Gabor}
T.~Lee.
\newblock Image representation using 2d gabor wavelets.
\newblock {\em TPAMI}, 18(10):959--971, 1996.

\bibitem{eth80}
B.~Leibe and B.~Schiele.
\newblock Analyzing appearance and contour based methods for object
  categorization.
\newblock In {\em CVPR}, 2003.

\bibitem{kle}
P.~Li, Q.~Wang, W.~Zuo, and L.~Zhang.
\newblock Log-euclidean kernels for sparse representation and dictionary
  learning.
\newblock In {\em ICCV}, 2013.

\bibitem{LADF}
Z.~Li, S.~Chang, F.~Liang, T.~Huang, L.~Cao, and J.~Smith.
\newblock Learning locally-adaptive decision functions for person verification.
\newblock In {\em CVPR}, 2013.

\bibitem{SD}
Z.~Liao, J.~Rock, Y.~Wang, and D.~Forsyth.
\newblock Non-parametric filtering for geometric detail extraction and material
  representation.
\newblock In {\em CVPR}, 2013.

\bibitem{eBiCov}
B.~Ma, Y.~Su, and F.~Jurie.
\newblock Bicov: a novel image representation for person re-identification and
  face verification.
\newblock In {\em BMVC}, 2012.

\bibitem{ARDataset}
A.~Martinez and R.~Benavente.
\newblock The ar face database.
\newblock In {\em CVC Technical Report 24}, 1998.

\bibitem{PCCA}
A.~Mignon and F.~Jurie.
\newblock Pcca: A new approach for distance learning from sparse pairwise
  constraints.
\newblock In {\em CVPR}, 2012.

\bibitem{GaborFace}
Y.~Pang, Y.~Yuan, and X.~Li.
\newblock Gabor-based region covariance matrices for face recognition.
\newblock {\em Circuits and Systems for Video Technology, IEEE Transactions
  on}, 18(7):989--993, 2008.

\bibitem{LF}
S.~Pedagadi, J.~Orwell, S.~Velastin, and B.~Boghossian.
\newblock Local fisher discriminant analysis for pedestrian re-identification.
\newblock In {\em CVPR}, 2013.

\bibitem{RFramework}
X.~Pennec, P.~Fillard, and N.~Ayache.
\newblock A riemannian framework for tensor computing.
\newblock {\em ICJV}, 66(1):41--66, 2006.

\bibitem{sphere_parameterization}
J.~Pinheiro and D.~Bates.
\newblock Unconstrained parameterizations for variance-covariance matrices.
\newblock {\em Statistics and Computing}, 6:289--296, 1996.

\bibitem{FastRegCov}
F.~Porikli and O.~Tuzel.
\newblock Fast construction of covariance matrices for arbitrary size image
  windows.
\newblock In {\em ICIP}, 2006.

\bibitem{BrodatzDataset}
T.~Randen and J.~Husoy.
\newblock Filtering for texture classification: a comparative study.
\newblock {\em TPAMI}, 21(4):291--310, 1999.

\bibitem{parameterization}
F.~Rapisarda, D.~Brigo, and F.~Mercurio.
\newblock Parameterizing correlations: a geometric interpretation.
\newblock {\em IMA J Management Math}, 18(1):55--73, 2007.

\bibitem{TSC}
R.~Sivalingam, D.~Boley, V.~Morellas, and N.~Papanikolopoulos.
\newblock Tensor sparse coding for region covariances.
\newblock In {\em ECCV}, 2010.

\bibitem{TSCwD}
R.~Sivalingam, D.~Boley, V.~Morellas, and N.~Papanikolopoulos.
\newblock Positive definite dictionary learning for region covariances.
\newblock In {\em ICCV}, 2011.

\bibitem{RegCov}
O.~Tuzel, F.~Porikli, and P.~Meer.
\newblock Region covariance: A fast descriptor for detection and
  classification.
\newblock In {\em ECCV}, 2006.

\bibitem{HumanD}
O.~Tuzel, F.~Porikli, and P.~Meer.
\newblock Human detection via classification on riemannian manifolds.
\newblock In {\em CVPR}, 2007.

\bibitem{PedD}
O.~Tuzel, F.~Porikli, and P.~Meer.
\newblock Pedestrian detection via classification on riemannian manifolds.
\newblock {\em TPAMI}, 30(10):1713--1727, 2008.

\bibitem{ColorName}
J.~van~de Weijer, C.~Schmid, J.~Verbeek, and D.~Larlus.
\newblock Learning color names for real-world applications.
\newblock {\em TIP}, 18(7):1512--1523, 2009.

\bibitem{CDL}
R.~Wang, H.~Guo, L.~Davis, and Q.~Dai.
\newblock Covariance discriminative learning: A natural and efficient approach
  to image set classification.
\newblock In {\em CVPR}, 2012.

\bibitem{logE-SR}
C.~Yuan, W.~Hu, X.~Li, S.~Maybank, and G.~Luo.
\newblock Human action recognition under log-euclidean riemannian metric.
\newblock In {\em ACCV}, 2009.

\bibitem{SDL}
Y.~Zhang, Z.~Jiang, and L.~Davis.
\newblock Discriminative tensor sparse coding for image classification.
\newblock In {\em BMVC}, 2013.

\bibitem{SalMatch}
R.~Zhao, W.~Ouyang, and X.~Wang.
\newblock Person re-identification by salience matching.
\newblock In {\em ICCV}, 2013.

\bibitem{eSDC}
R.~Zhao, W.~Ouyang, and X.~Wang.
\newblock Unsupervised salience learning for person re-identification.
\newblock In {\em CVPR}, 2013.

\bibitem{MidFilter}
R.~Zhao, W.~Ouyang, and X.~Wang.
\newblock Learning mid-level filters for person re-identification.
\newblock In {\em CVPR}, 2014.

\end{thebibliography}
}

\end{document}